\documentclass[letterpaper, 10 pt, conference]{ieeeconf}  

\IEEEoverridecommandlockouts                              

\usepackage{graphicx}
\usepackage{amsmath} 
\usepackage{fancyhdr} 
\usepackage{hyperref}

\newcommand{\eref}[1]{(\ref{#1})} 
\newcommand{\fref}[1]{Fig.~\ref{#1}} 
\newcommand{\sref}[1]{Section~\ref{#1}} 
\newcommand{\tref}[1]{Table~\ref{#1}} 
\newcommand{\boldSubSec}[1]{\textbf{#1:}}

\title{\LARGE \bf
High-Performance Reinforcement Learning on Spot: \\ Optimizing Simulation Parameters with Distributional Measures
}

\author{
    A.J. Miller$^{1,2}$,
    Fangzhou Yu$^{1}$,
    Michael Brauckmann$^{1}$,
    and Farbod Farshidian$^{1}$
    \thanks{$^{1}$
        RAI Institute,
        Cambridge, MA 02139, USA:
        {\tt\small amiller@rai-inst.com, ffarshidian@rai-inst.com}
    }
    \thanks{$^{2}$
        Department of Electrical Engineering and Computer Science, 
        MIT, Cambridge, MA 02139, USA
    }
}

\begin{document}

\maketitle
\thispagestyle{empty}
\pagestyle{empty}


\begin{abstract}

This work presents an overview of the technical details behind a high-performance reinforcement learning policy deployment with the \emph{Spot RL Researcher Development Kit} for low-level motor access on Boston Dynamics' Spot.
This represents the first public demonstration of an end-to-end reinforcement learning policy deployed on Spot hardware with training code publicly available through NVIDIA Isaac Lab and deployment code available through Boston Dynamics.
We utilize Wasserstein Distance and Maximum Mean Discrepancy to quantify the distributional dissimilarity of data collected on hardware and in simulation to measure our sim-to-real gap.
We use these measures as a scoring function for the Covariance Matrix Adaptation Evolution Strategy to optimize simulated parameters that are unknown or difficult to measure from Spot.
Our procedure for modeling and training produces high-quality reinforcement learning policies capable of multiple gaits, including a flight phase.
We deploy policies capable of over 5.2m/s locomotion, more than triple Spot's default controller maximum speed, robustness to slippery surfaces, disturbance rejection, and overall agility previously unseen on Spot.
We detail our method and release our code to support future work on Spot with the low-level API.

\end{abstract}

\section{INTRODUCTION}

Boston Dynamics’ Spot \cite{spot} is known the world over for opening doors \cite{spot_open_door}, working in factories \cite{spot_work_3}, and its many dances \cite{spot_dance_1}.
It has captured the curiosity of the public and the imagination of roboticists about what legged robots can begin to look like in everyday life.
With the release of its SDK \cite{spot_sdk} and the commercial launch of the hardware platform, Spot took some of the first steps for a highly articulated robot toward successful real-world commercial applications.
In another step earlier this year, Boston Dynamics unveiled their first inclusion of reinforcement learning (RL) techniques in their control stack \cite{spot_rl}.

In continuation of these advancements, we present in this work the training and deployment of the first fully learned control policy on Spot hardware, describe a sim-to-real gap quantification procedure using only on-board sensing, and provide a simulation parameter optimization process based on this quantification.
We demonstrate the capability of our procedure by training a policy to push the limits of Spot.
We show new speed, agility, and robustness with our end-to-end policy.
We describe our modeling, training, and deployment as an example for researchers and engineers to train and deploy their own RL policies on Spot.
\\

\noindent Our contributions are the following:
\begin{enumerate}

\item An evaluation procedure for measuring sim-to-real gap inspired by generative learning techniques and an optimization procedure for selecting simulator parameters
\item The first end-to-end RL control policy on Spot hardware and open-source training code
\item Demonstrations of the extended capabilities of our control policy’s robustness and agility, including a more than triple maximum forward velocity over the default Spot controller. Videos of these results can be found online. \footnote{\href{https://www.youtube.com/watch?v=BolpYgX36DA}{https://www.youtube.com/watch?v=BolpYgX36DA}}

\end{enumerate}

\begin{figure}[t]
  \centering
  \includegraphics[width=\columnwidth]{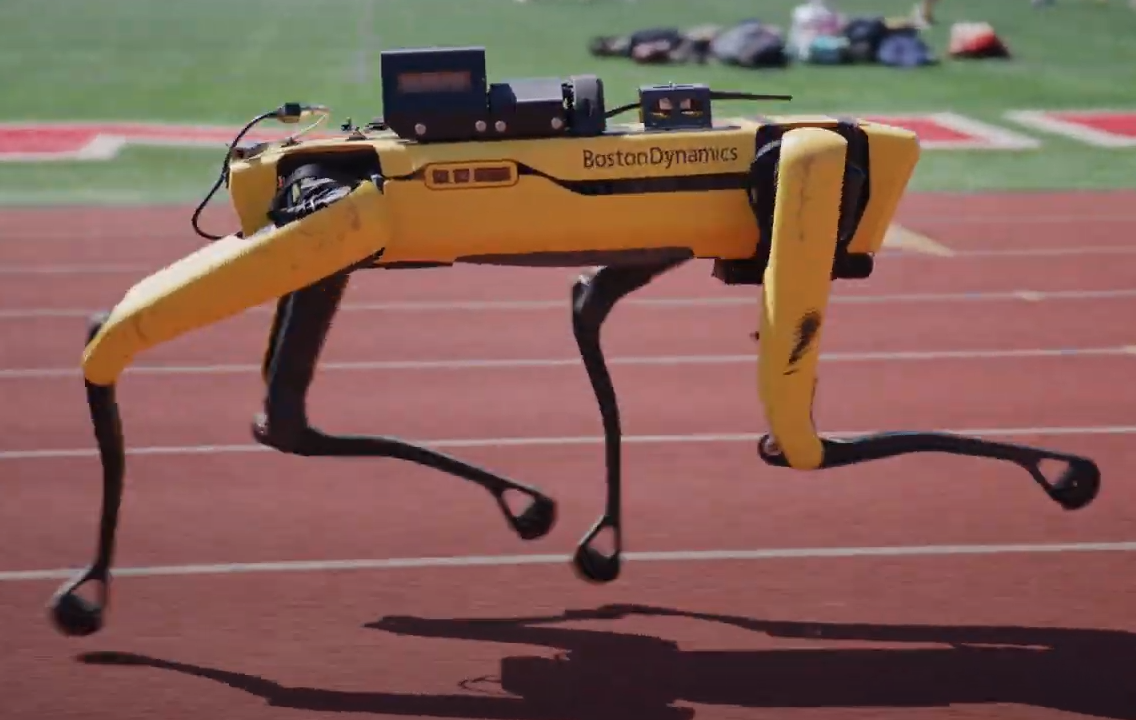}
  \caption{Spot galloping in flight on a flat track with our policy at 5.2m/s.}
  \label{spot_cover}
\end{figure}

\section{RELATED WORKS}

The last decade has developed many highly articulated fully electric quadruped robots like Spot.
These have included MIT’s Cheetah 3 \cite{bledt2018cheetah}, IIT’s HyQ \cite{semini2011design}, and ANYbotic’s ANYmal \cite{hutter2016anymal}.
Traditionally these systems have been controlled by model-based optimization schemes such as Model Predictive Control (MPC) \cite{di2018dynamic, farshidian2017real} and Whole Body Control (WBC) \cite{bellicoso2017dynamic} but in recent years, RL has demonstrated itself as a reliable, performant and robust tool in developing policies for the real world \cite{hwangbo2019learning, margolis2024rapid, hoeller2024anymal} while leveraging the advantages of simulation parallelization \cite{rudin2022learning}.
RL enables concurrent training with specialized networks \cite{ji2022concurrent}, adaptation during deployment \cite{kumar2021rma}, and new dynamism on existing platforms including higher speeds \cite{margolis2024rapid, shin2023actuator} and entirely new behaviors like parkour \cite{hoeller2024anymal}.

Policy training in the simulation enables fast data collection, but simulation rollouts may not accurately match hardware.
Sim-to-real differences can dramatically impede performance \cite{muratore2022robot}\cite{xie2018feedback} and many methods have been proposed to minimize this gap.
Domain randomization enables the policy to capture the hardware within the learned distribution \cite{exarchos2021policy, li2021reinforcement} but may cause degradation in the performance and precision of the policy.
Actuator networks \cite{hwangbo2019learning} replicate actuator input/output behaviors and show great performance but require reliable joint torque measurements.
Others have forgone improving the simulation altogether and focused on tuning the policy during deployment \cite{kumar2021rma}.

System identification approaches instead search for better values for model parameters \cite{shin2023actuator, kaspar2020sim2real} from the evaluation of hardware data.
A naive approach compares trajectory rollouts directly; this is difficult, however, because errors compound and result in substantial drift.
Single-step updates are often used instead and evaluated with state-based methods such as least squares \cite{an1985estimation, lee2019geometric}.
This requires accurate simulation resets to match the hardware state, but this is not always feasible.
For example, states like contact need body height information that isn’t directly measurable from onboard sensing.
With modern simulators, we can instead easily generate large amounts of data and evaluate them in aggregate.

Generative learning provides examples of evaluating performance like this and many solutions focus on the data distribution \cite{borji2019pros}.
Specifically, Wasserstein distance \cite{ruschendorf1985wasserstein} and Maximum Mean Discrepancy (MMD) \cite{dziugaite2015training} have been used with great success to evaluate the performance of Generative Adversarial Networks (GANs) by comparing features of the distribution and are indifferent to compounding error like in time-series data.
Applications of these measures in robotics include multi-modal behaviors with GAN structures \cite{peng2021amp}, imitation learning using GAN-style losses \cite{li2023learning}, and transfer learning of exteroceptive information \cite{rahman2021knowledge}.
These methods of scoring performance based on distributional characteristics instead of state-based evolution of dynamics leverage the advantages of modern simulators without the onerous requirement that the initial states match.


\section{METHODS}

Here we detail our methods to model Spot and train our robust high-speed RL policies.
We explain our modeling of both parameters known directly from measurements or the specifications of Spot as well as parameter estimates for joint friction and the torque-speed limit profiles using our scoring methodology.
We collect state rollouts on hardware and in simulation and use Wasserstein Distance and MMD to quantify distributional similarity.
We optimize our similarity score using the Covariance Matrix Adaptation Evolution Strategy (CMA-ES) on the parameter estimates to improve our model.
We train a high-speed locomotive policy using our optimized Spot model and describe our training procedure.

\subsection{Modeling}

Modeling the actuators as best as possible is key so that the policy remains in distribution when deployed on the platform.
In the case of Spot, the motor controllers use torque-sensing feedback to achieve the desired torque in the motor output.
This alleviates some common modeling and randomization including accounting for gear ratios or unreliable actuators.
However, we still need to account for communication delay within the low-level API, quantify state estimation noise for the policy observations, include torque limits, and approximate frictional effects within each joint.

\begin{figure}[t]
  \centering
  \includegraphics[width=\columnwidth]{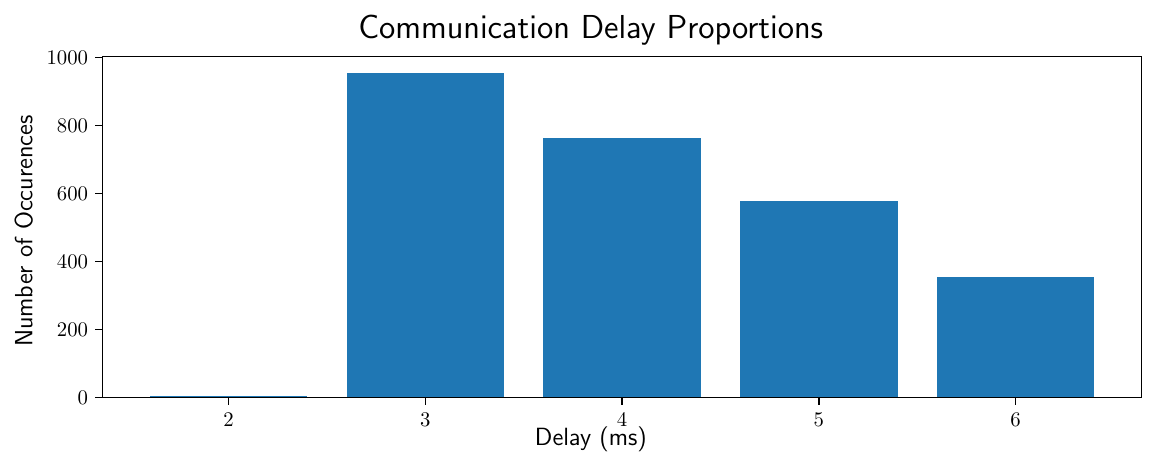}
  \caption{Histogram plot of the measured communication delay of Google Remote Procedure Call messages sent to the Spot API. This is used to estimate the delay in torque application we use in the simulated actuators.}
  \label{delay}
\end{figure}

\boldSubSec{Communication Delay}
We communicate with the motors using Google Remote Procedure Calls (gRPC) \cite{grpc}.
We quantified communication delay by measuring the time to send a command and receive application confirmation.
We collected sample times in milliseconds and show the results in \fref{delay}.
To model the delay within our simulation, we buffer the action outputs of the network for 5ms.
Delaying action application the total delay is identical to delaying both the observations and actions.

\begin{figure}[t]
  \centering
  \includegraphics[width=\columnwidth]{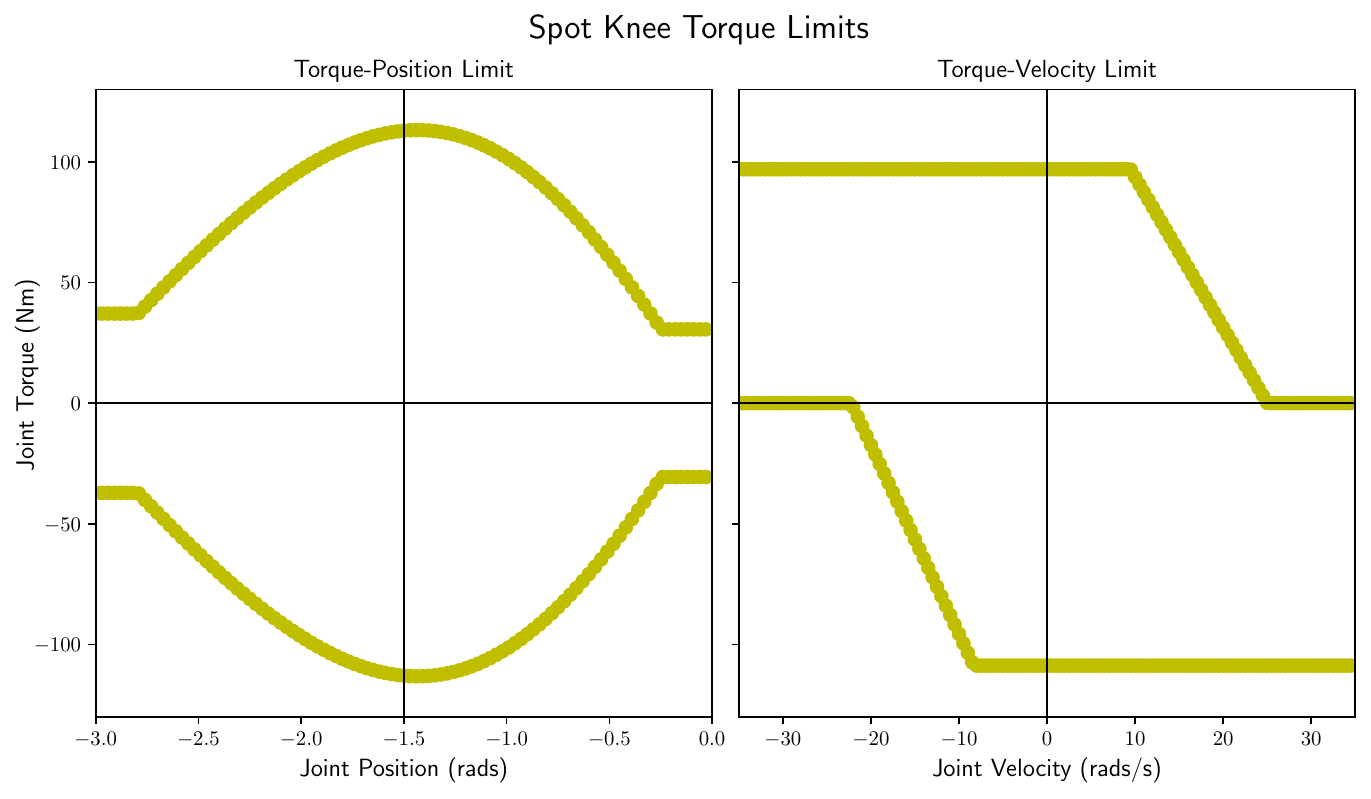}
  \caption{Torque-position (left) and torque-speed (right) limits of the remotized knee actuator scatter plotted into two dimensions. These values are used to cap the applied torque in the simulated joints to better match hardware limitations.}
  \label{torque_limits}
\end{figure}

\boldSubSec{Actuators}
Each leg of Spot has three actuators; from the base towards the foot: hip$_x$, hip$_y$, and knee.
The motor packages are identical for each, but the knees also have a remotized lead-screw joint.
This joint has a higher friction coefficient, and as expected we found the optimized knee friction values to be higher than for the hips \tref{tab_cmaes_output}.
We model torque limits for the joints to cap the torque available for the policy to request.
The torque limits of the hip joints are set to flat maximum values.
The torque limits of the remotized knees depend on the joint's position and velocity.
We model these limits as shown in \fref{torque_limits}.
We set the torque-position limits using the Spot API documentation \cite{spot_sdk} and torque-speed limits from our method described in \sref{optimization}.
The limit is computed as the minimum of the torque-position and torque-speed limits.

\boldSubSec{Observation Noise}\label{obs_noise}
We reduce the observation sim-to-real gap using data-driven methods to construct a Gaussian noise model.
Although empirical results indicate training with corrupted observations is not necessary for good sim-to-real behavior on Spot, we found noise modeling results in more robust policies and accelerates training by injecting more exploration into the policy rollout.
For every observation feature \( o_t \), we apply a white noise model by sampling a Gaussian distribution: $\mathbf{n}_t \sim \mathcal{N}(0, \sigma_t)$, for the corrupted observation $\hat{\mathbf{o}}_t = \mathbf{o}_t + \mathbf{n}_t$.
The values for $\sigma_t$ are derived from data collected from the Spot state estimator throughout a conservative walking gait and logged at the control frequency $f_s$.
The raw data is transformed into the frequency domain using a Fast Fourier Transform (FFT).
We then normalize the FFT by multiplying by the frequency $f$ such that the energy content of the noise signal is constant across all frequencies \cite{mignard2005nyquist}.
The value for $\sigma_t$ is estimated by computing the average signal power of the normalized FFT over the right half of the frequency domain $[\frac{f_s}{4}, \frac{f_s}{2}]$ where $\frac{f_s}{2}$ is the Nyquist frequency.


\subsection{Similarity Scoring of Simulator Performance}


When training in a simulator, there will be inevitable mismatches between the simulated robot and the real hardware.
Minimizing the sim-to-real gap is a necessity for high-performance hardware results, and many methods have been proposed in the literature to mitigate this challenge \cite{ji2022concurrent, kumar2021rma, tan2016simulation, tan2018sim, james2019sim, buchanan2022learning}.
In one extreme, a first principles approach may attempt to estimate uncertain physical parameters such as the actuator model \cite{hutter2013starleth}; however, the complexity can be quite high and quickly become arduous to replicate on new systems.
On the other extreme, a black-box learned approach such as the actuator net \cite{hwangbo2019learning}, may obfuscate the full details.
Additionally, the learned approach relies on output torque measurements, which many systems do not have, such as robots with pseudo-direct-drive actuators \cite{bledt2018cheetah, katz2019mini}, or may require invasive procedures to disassemble and take measurements, which may not be simple or possible.

We balance these considerations with a gray-box parameterization.
Features may include joint friction, actuator armature, torque-speed curve limits, or power supply dependencies on temperature or charge.
The limitations we observed in our original deployments were lower measured torque than expected, slower swing cycle re-circulation, and joint position overshooting.
We focused on joint friction and torque-speed curve limits as our hyperparameters.

Regardless of the hyperparameter space, defining evaluation metrics for quantifying the sim-to-real gap is another daunting task.
We were inspired by the techniques used in generative learning, where the alignment of qualitatively ``good" desired behavior and quantitative formulations used to train is very difficult.
A common approach in generative learning is to use statistical measures to evaluate the distributional differences between the target and sample data \cite{borji2019pros}.
We see an analogy where the execution of our policy in simulation acts like a generator and the real data is the execution of our policy on hardware.
Similarly, we aim to minimize distributional differences between simulated rollouts and pre-collected hardware for a given policy.


We utilized two distribution measurement methods widely adopted in generative applications: Wasserstein Distance \cite{ruschendorf1985wasserstein} and MMD \cite{dziugaite2015training}.
These measures are known for their high discriminability, boundedness, ease of implementation, and lack of dependency on pre-trained models \cite{borji2019pros}, which make them prime candidates for our application.
Wasserstein distance, also known as Earth Mover’s distance, measures the minimum distance needed to convert one probability distribution into another so that the shapes of the distributions match.
MMD measures the difference in the means of the distributions projected into an embedded higher dimensional space.
We used both measures to quantify the dissimilarity between hardware and simulation data.

\begin{figure}[t]
  \centering
  \includegraphics[width=\columnwidth]{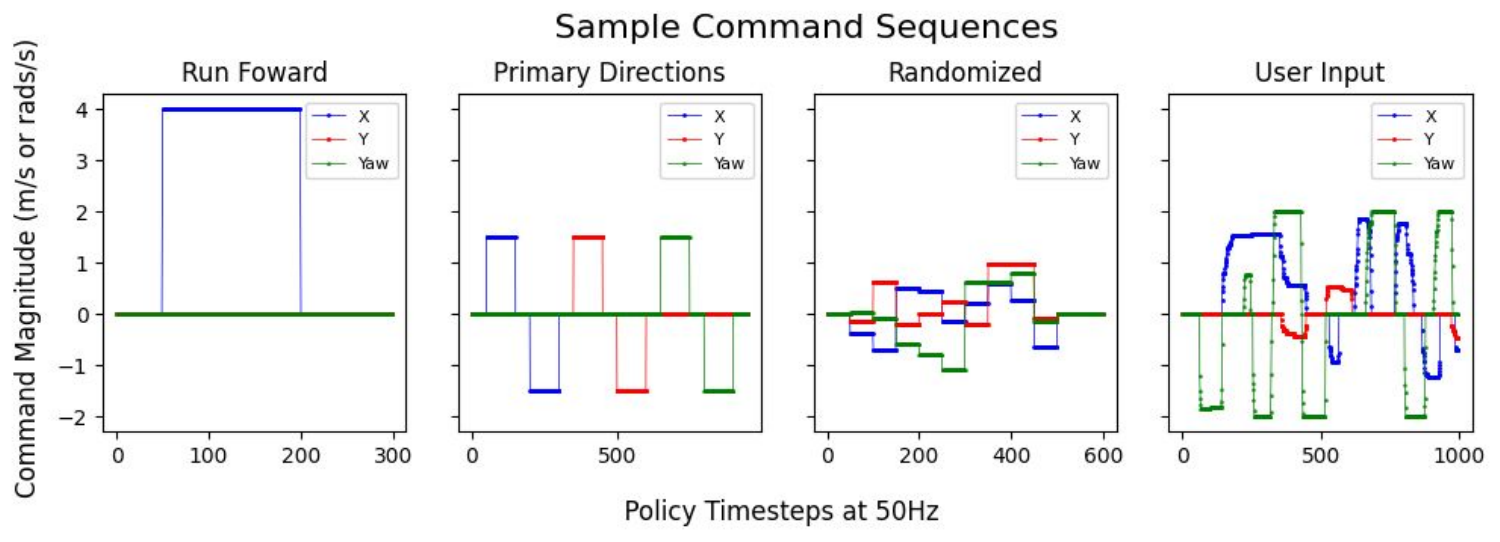}
  \caption{Command sequence used to generate test data for evaluation of hardware and simulation similarity. This includes running at 4m/s, moving 1.5m/s in all six command directions, randomized, and user commands.}
  \label{command_plots}
\end{figure}

A final consideration in using the evaluation metrics is the feature space used to represent the samples.
We chose measured joint position, velocity, and policy action\footnote{To normalize the feature space, we scaled the joint position by a proportional joint gain $k_p$, joint velocity by damping joint gain $k_d$, and the policy action by action scaling factor $\sigma_a$ and $k_p$.}.
For data collection, we used a series of scripted commands to gather data.
We used four different command sequences, shown in \fref{command_plots}.
The sequences included running forward, sampling the six commandable directions at moderate velocity, randomly changing commands, and a series of user commands.
We sampled each 5 times on hardware and repeated the commands in the simulation.
We rolled out multiple simulated robots with each command sequence and calculated the measure distance between hardware/simulation pairs with the same command sequence.
We aggregated the variations together by taking the average of the distance measure.




\subsection{Hyperparameter Optimization} \label{optimization}

With measures to quantify the similarity of policy performance, we can devise an optimization on the multi-dimensional hyperparameter space to reduce the sim-to-real gap.
We first train a policy in simulation with the default hyperparameter values.
Next, we deploy this policy on hardware and log the real data.
Based on the real data, we perform a hyperparameter optimization.
We then retrain a new policy using the updated hyperparameter values.
Some fine-tuning of the reward functions may be required at this step.
Finally, we deploy this newly trained policy on hardware.
In practice, we found that a single iteration of this process was sufficient to find performant simulation hyperparameters but the process could be repeated if desired.

We chose joint friction and torque-speed limits with a simplified model of 8 hyperparameters to optimize.
The model includes two friction parameters with $\text{hip}_x$ and $\text{hip}_y$ set jointly and the remotized knee friction set separately.
We created a 6-parameter simplification of the torque-speed curve with maximum and minimum torques, maximum and minimum speeds, and the torque-speed trade-off in quadrants 1 and 3 through the intersection between the torque and speed limits.
A torque-speed curve limit can be seen in \fref{torque_limits}.

For our optimization, we used the CMA-ES \cite{auger2012tutorial} as a gradient-free sampling method to set the hyperparameters and minimize distributional measure error.
We selected CMA-ES because of its successful applications in hardware design \cite{chen2020hardware, fadini2022simulation} and motion generation \cite{xie2018feedback, dehio2015multiple, li2015kicking}.
The algorithm works well in a non-smooth optimization landscape and scales well for tens to one hundred search parameters.

\begin{table}[t]
    \centering
    \caption{Joint friction, torque, and speed limit parameters optimized by our method described in \sref{optimization} and used in training.}
    \begin{tabular}{lcc}
    Parameter & Values & \\
    \hline \hline \\
    Friction & Hips: 0.008Nm & Knees: 0.180Nm \\
    \hline \\
    Torque Limits & Max: 97.00Nm & Min: -108.79Nm \\
    \hline \\
    Speed Limits & Max: 25.03rad/s & Min: -22.22rad/s \\
    \hline \\
    \vtop{\hbox{\strut Torque-Speed}\hbox{Intersect}} & Max: 9.48rad/s & Min: -8.32rad/s \\ \\
    \hline \hline \vspace{-8pt}
    \end{tabular}
    \label{tab_cmaes_output}
\end{table}

We ran CMA-ES for 100 iterations with a sample population size of 10.
For each iteration, we run the simulator with a new hyperparameter sample, collect simulated data based on the command generation sequence, and evaluate using the measures to score the similarity to real data.
Afterward, we selected the parameters shown in \tref{tab_cmaes_output} and iterated the learning process by retraining a new policy.



\subsection{Training}

\begin{figure*}[t]
  \centering
  \includegraphics[width=\textwidth]{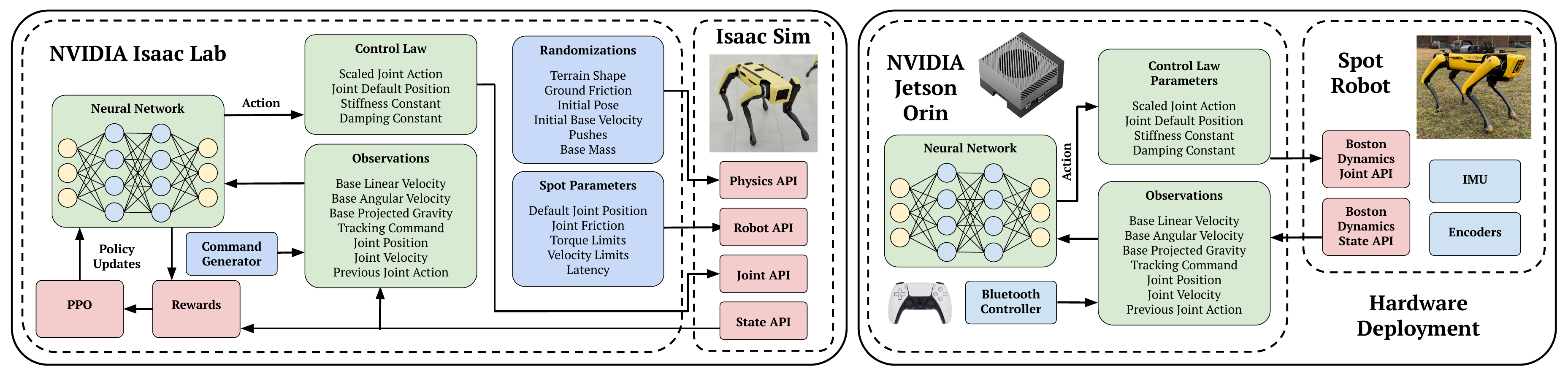}
  \caption{Visual overview of the training and deployment pipeline. On the left, is an approximation of how the policy is trained in simulation, and on the right, is how it's deployed on hardware. Isaac Lab is the simulation environment, provides the physics engine, and employs our modeling for training. We deploy our policies on Spot using an NVIDIA Jetson Orin that communicates with the Boston Dynamics API for state estimates and sends motor control law values with the \emph{Spot RL Researcher Development Kit} to command the motors.}
  \label{diagram}
\end{figure*}

We trained our policies using the NVIDIA Isaac Lab \cite{mittal2023orbit} framework built on NVIDIA Isaac Sim \cite{isaacsim} and with learning algorithms from RSL\_RL \cite{rslrl}.
Isaac Lab provides a general RL training environment to model and simulate the Spot, interfaces with Isaac Sim for observations, actions, and rewards, and parallelizes data collection and neural network training.
We train a policy using the Actor-Critic \cite{grondman2012survey} formulation and updated with Proximal Policy Optimization (PPO) \cite{schulman2017proximal}.
An overview of the training pipeline can be seen in \fref{diagram}.

\boldSubSec{Actions}
The policy is sampled at 50Hz and the actions are converted into motor torques $\tau$ by the PD control law as defined in \eref{control_law}.
The control law calculates the desired torque from the set points and gains and is then applied at the simulated joints at 200Hz.

\begin{equation}\label{control_law}
    \tau = k_p (\sigma_a a + q_d - q) - k_d\dot{q},
\end{equation}
$k_p$ and $k_d$ are the stiffness and damping gains for the actuators.
$k_p$ is 60.0 N$\cdot$m$\cdot$rad$^{-1}$ and $k_d$ is 1.5 N$\cdot$m$\cdot$s$\cdot$rad$^{-1}$ for training and deployment on all 12 Spot actuators.
The policy output action $a$ is multiplied by a scaling factor $\sigma_a = 0.2$. The action is represented as a joint position set-point offset and is summed with the default joint position $q_d$ minus the measured joint position $q$.
The negative of measured joint velocity $\dot{q}$ damps towards zero.


\boldSubSec{Observations}
The observations for the neural network are described in \eref{obs_eq}.
They are collected as the simulated states of the robot for training and from the state estimator when on hardware.

\begin{equation}\label{obs_eq}
    o = \big( v_{xyz}, \omega_{xyz}, g, {cmd}, q, \dot{q}, a_{t-1} \big),
\end{equation}
$v_{xyz}$ is the body relative base linear velocity where $x$ is longitudinal, $y$ is lateral, and $z$ is vertical axes.
$\omega_{xyz}$ is the base angular velocity.
$g$ is the normalized gravity vector projection in the body frame.
This acts as a proxy for the base orientation.
${cmd}$ is the velocity tracking command of the robot consisting of base longitudinal $V_x$, lateral $V_y$, and yaw $\omega_z$.
Commands were sampled in the range $V_{cmd,x}{\sim}U(\text{-}2.0, 5.5)$m/s, $V_{cmd,y}{\sim}U(\text{-}1.5, 1.5)$m/s, and $\omega_{cmd,z}{\sim}U(\text{-}2.0, 2.0)$rad/s.
$q$ are the joint angles, $\dot{q}$ are the joint velocities, and $a_{t-1}$ are the previous network actions.


\boldSubSec{Rewards}
The task rewards include tracking the base linear ($r_{v_{xy}}$) and base angular ($r_{\omega_{z}}$) velocity tracking in the three commanded axes, and a penalty for base velocity in the three non-commanded axes ($r_{v_z\omega_{xy}}$).
The style rewards include the time each foot spends in the air versus in contact ($r_{at}$), penalizing the variance of the total air and contact time of the previous step between the feet ($r_{atv}$), the vertical foot clearance from the ground during swing and stance ($r_{fc}$), penalizing the impact velocity of the foot when transitioning from swing to stance ($r_{fi}$), penalizing foot slippage during stance ($r_{fs}$), penalizing low distance between the feet ($r_{fd}$), penalizing deviation from flat orientation ($r_{\phi}$) relative to the ground plane, and encouraging swing/stance synchronization for the desired gait ($r_g$).
The regularization rewards include penalizing large changes in the current action from the previous action ($r_{as}$), penalizing deviation from the default joint position ($r_{q}$), penalizing joint velocity ($r_{\dot{q}}$), penalizing joint acceleration ($r_{\ddot{q}}$), and penalizing joint torque ($r_{\tau}$).

\boldSubSec{Environment}
We randomized the terrain, ground friction, base mass, initial body position $X_0$ and velocity $V_0$, and initial joint position $q_0$ and velocity $\dot{q_0}$.
We randomized a low poly noisy terrain to deviate from flat.
The foot-ground friction was randomized between $U(0.3, 1.0)$ for static and $U(0.3, 0.8)$ for dynamic friction.
Additional mass of the base $\delta_{m,\text{base}}{\sim}U(\text{-}2.5, 2.5)$, initial body position $X_{xy,0}{\sim}U(\text{-}0.5, 0.5)$, $\phi_{z,0}{\sim}U(\text{-}\pi, \pi)$ and body velocity $V_{x,0}{\sim}U(\text{-}1.5, 1.5)$, $V_{y,0}{\sim}U(\text{-}1.0, 1.0)$, $V_{z,0}{\sim}U(\text{-}0.5, 0.5)$, $\omega_{x,0}{\sim}U(\text{-}0.7, 0.7)$, $\omega_{y,0}{\sim}U(\text{-}0.7, 0.7)$, $\omega_{z,0}{\sim}U(\text{-}1.0, 1.0)$, deviation from initial joint position $\delta_{q,0}{\sim}U(\text{-}0.2, 0.2)$ and velocity $\dot{q_0}{\sim}U(\text{-}2.5, 2.5)$ were uniformly sampled in these ranges and fixed at the beginning of the episode.
We randomly pushed the robot by modifying its base velocity in the range $V_{xy}{\sim}U(\text{-}0.5, 0.5)$.
We did not randomize joint friction, stiffness, damping, torque limits, ground restitution, action scale, limb mass, or apply forces on the body during training.

\boldSubSec{Termination}
We terminate an episode and reset the agent under three conditions: if the agent reaches the episode timeout maximum of 20 seconds; if the agent touches the ground with its base or upper or lower leg; or if the agent goes beyond the terrain boundary.
We considered this out-of-bound condition similar to a timeout as it happens when the robot successfully runs to the terrain boundary.
We found this to be a reasonable trade-off instead of increasing the terrain size due to its impact on simulation performance.


\begin{figure}[t!]
  \centering
  \includegraphics[width=\columnwidth]{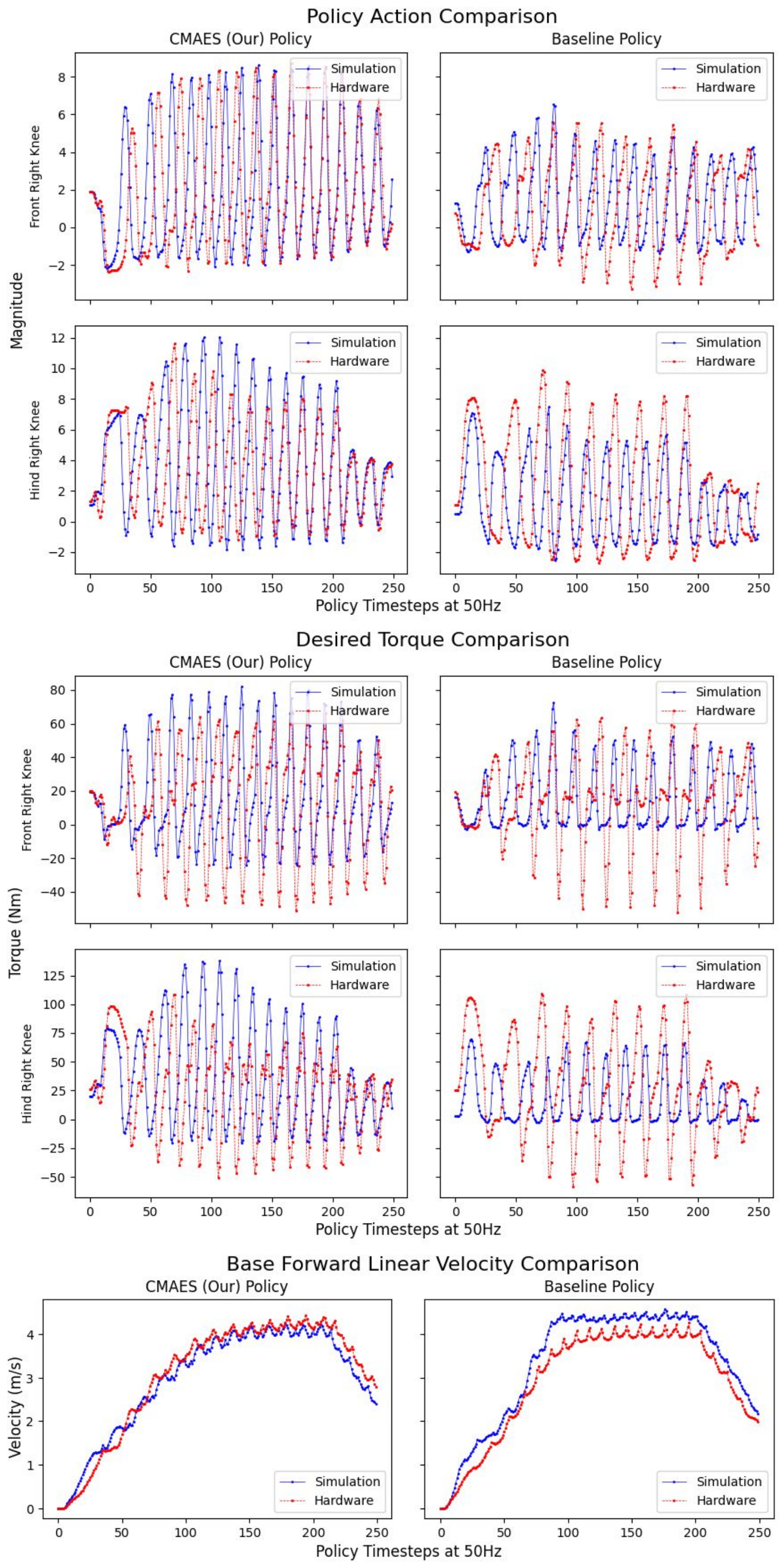}
  \caption{Comparison of knee joint action and desired torque during a commanded 4.0m/s acceleration on hardware and in simulation comparing the performance of our CMA-ES modeling trained policy versus a baseline policy. The improvements are apparent in the better base velocity matching and amplitude and frequency matching of joints during the gait cycle.}
  \label{comp_curves}
\end{figure}

\subsection{Deployment}

In the Spot API, we communicate with the state estimator and motors via gRPC messages.
We query the API for the state estimates and input the concatenated states into the trained neural network, represented in runtime ONNX.
The network output is used as a joint position set-point offset.
This offset is sent with the default joint positions, joint velocity set-points (set to 0.0rad/s), stiffness ($k_p =$ 60.0 N$\cdot$m$\cdot$rad$^{-1}$), and damping ($k_d =$ 1.5 N$\cdot$m$\cdot$s$\cdot$rad$^{-1}$) in a gRPC message to the low-level motor control API.
The desired joint torques are calculated from these values and updated using measured torque feedback.

\section{RESULTS}




In this work, we pushed Spot in ways previously unseen to better understand the limitations of the platform.
RL is a powerful tool for this and accurate modeling is important for best results.
Our method produces the necessary modeling without disassembling the hardware to achieve our desired performance.
We demonstrate a more than tripling of the maximum forward velocity, new gaits including a flight phase, and robustness to disturbances on slippery ground surfaces.
We showcase these abilities and results in the supplementary video and project website.

\subsection{CMA-ES Optimization}

We ran our CMA-ES optimization using a weighted average of the measure scores for each of the four command sequences.
We initialized the values for actuator frictions to 0.0Nm, torque limits to +/-70Nm, torque speed limits to +/-20rad/s, and torque-speed intersects at +/-0.0rad/s.
The optimization resulted in the values of the actuator friction and torque-speed curves shown in \tref{tab_cmaes_output}.
We minimize the distributional differences between the joint positions, joint velocities, and actions of the policy.
The values are normalized by their respective control law scaling constants.
Importantly, simulated data does not include sensor noise.

We used these values to retrain our policy where only minor weight value adjustments were made to maintain behavior quality.
We found a single iteration of the algorithm sufficient to markedly improve performance and increase maximum linear velocity from 3.8m/s to 5.2m/s on hardware.
These improvements also enabled flight phase gaits as seen in \fref{spot_cover}, which previously failed on hardware.


\subsection{Comparison to Baseline on Hardware}

Empirical results of our method's improvements in minimizing sim-to-real differences are shown in \fref{comp_curves}.
For these results, we deployed a baseline policy trained without our modeling and a policy trained with the CMA-ES modeling.
Both policies were commanded from a standstill to run forward at an increasing velocity until 4.0m/s.
Rollouts were collected on hardware and in simulation using the same settings as training.
We plotted the actions and desired torques for the front and hind right knees as these joints are out of phase during the trot gait, and the knees required the most modeling with our method.

As shown in the figures, the joint actions and torque more closely match in simulation and on hardware for our policy than the baseline.
This can be seen in the magnitude and frequency of the swing and the consistency of the motion.
Notably, both policies struggle to produce negative torque during swing leg recirculation resulting in a slower swing and larger error that increases the desired torque value.
This is due to power distribution limits not being included in our modeling, however, our policy is more resistant to these effects as seen in the greater stability in the joint actions. 
This causes a larger difference in the swing leg time between the baseline simulation and the hardware reducing overall performance.

Importantly our better joint behavior matching results in better forward base linear velocity matching.
The divergences seen in the baseline policy deployment are critically detrimental to the overall performance of the controller.
The baseline policy is too out of distribution on the hardware and results in nontrivial performance degradation which is substantially mitigated by our method.

\subsection{Robustness and High-Performance}

A valuable randomization for policy robustness is the ground friction coefficient.
Our policy testing predominantly used moderate to high friction coefficient surfaces.
To test our policy, we ran Spot on a low-friction surface created by applying soap and water on plexiglass plates.
We ran tests on this surface to show the policy adapting to sudden changes in friction and recovering from foot slips, including staying upright after a knee collided with the ground.

The primary demonstration of our method's advantages is in the high-speed performance of our policy.
We deployed on a rubberized outdoor running track to find the maximum velocity of our approach.
In our testing, we reached sustained running speeds of 5.2m/s in a flying trot gait; exceeding our baseline policy's 3.7m/s sustained maximum and far exceeding the default Spot controller's fastest forward velocity of 1.6m/s \cite{spot_specs}.
Our modeling of the noise in the state estimation, joint friction effects, and torque limits are necessary to reach this level of performance and validate the efficacy of our method in finding good parameter values for our simulated robot model.
These demonstrations can be found in the supplementary video.

\section{CONCLUSION}

In this work, we present the first end-to-end RL deployment on Spot, detail our modeling, training, and deployment procedure, and release our training code to support future research with the \emph{Spot RL Researcher Developer Kit}.
We provide this as a demonstration of our workflow using Spot low-level motor access.
We quantify and evaluate the sim-to-real gap of our simulator model to the real hardware using a Wasserstein Distance and MMD-based scoring mechanism and create an optimization procedure with CMA-ES to tune our simulator parameters for better performance.
Our process produces a high-quality Spot model of Spot and its actuators to enable greater speed, agility, and robustness never before seen on the platform.
While there is more to do in leveraging Spot's vision capabilities, improving the modeling in simulation, and pushing the hardware with even more ambitious behaviors, we believe this is a foundational first step in enabling broader future work.

\section*{ACKNOWLEDGMENT}

We extend our deep appreciation to multiple individuals for their valuable assistance.
We thank Boston Dynamics and Ben Schwilling for help with the Spot low-level API and in understanding the actuators of the robot.
We thank Surya Singh for his thoughts on modeling the state estimation sensor noise.
We thank Kyle Morgenstein for testing early versions of training and assisting with deployment on hardware.
We thank Emmanuel Panov and Joe St Germain for critical hardware support.
We thank Gabe Nelson and Al Rizzi for their insightful feedback and continuous willingness to help us better understand Spot.



\bibliographystyle{IEEEtran} 
\bibliography{refs.bib} 

\newpage

\addtolength{\textheight}{-12cm}  




\end{document}